\pdfoutput=1 

\newif\ifislncs
\newif\ifisieee

\islncsfalse
\isieeetrue

\ifislncs
	\documentclass{llncs} %
\fi

\ifisieee
	\documentclass{article}
	\usepackage{spconf}
\fi

\usepackage{vmargin} 
\setpapersize{A4}
\setmargins{3.5cm}
					 {1.5cm}
           {14.7cm}
           {23.42cm}
           {14pt}
           {1cm}
           {0pt}
           {2cm}

\usepackage{t1enc} 
\usepackage[latin1]{inputenc} 
\usepackage{times}		

\usepackage[pdftex]{graphicx} 
\usepackage[font=small,labelfont=bf,tableposition=top]{caption}
\usepackage{subcaption}
\usepackage{color} 

\DeclareCaptionLabelFormat{andtable}{#1~#2  \&  \tablename~\thetable}
\usepackage[fleqn]{amsmath}
\usepackage{amssymb}

\usepackage[switch]{lineno}

\usepackage[pdfborder=000,pdftex=true]{hyperref}

\ifisieee
\title{Multi-scale Visual Attention \& Saliency Modelling \\ with Decision Theory}
\name{Anh Cat Le Ngo\(^{1,2}\) \quad Li-Minn Ang\(^{3}\) \quad Guoping Qiu\(^{2}\) \quad Kah Phooi Seng\(^{4}\)}
\address{\(^{1}\) Faculty of Engineering, University of Nottingham, Malaysia Campus, Malaysia \\
		 \(^{2}\) School of Computer Science, University of Nottingham, Jubilee Campus, UK \\    
	     \(^{3}\) Centre for Communications Engineering Research, Edith Cowan University, Australia \\
	     \(^{4}\) Department of Computer Science \& Networked System, Sunway University, Malaysia}
\fi

\begin{document}
\linenumbers


\maketitle

\begin{abstract}

Bottom-up saliency, an early human visual processing, behaves like binary classification of interest and null hypothesis. Its discriminant
power, mutual information of image features and class distribution, is closely related to saliency value by the well-known centre-surround theory. As classification accuracy very much depends on window sizes, the discriminant saliency (power) varies according to sampling scales. Discriminating power estimation in multi-scales framework needs integrating with wavelet transformation and then estimating statistical discrepancy of two consecutive scales (centre-surround windows) by Hidden Markov Tree (HMT) model. Finally, multi-scale discriminant saliency (MDIS) maps are combined by the maximum information rule to synthesize a final saliency map. All MDIS maps are evaluated with standard quantitative tools (NSS,LCC,AUC) on N.Bruce's database with ground truth data as eye-tracking locations ; as well assessed qualitatively by visual examination of individual cases. For evaluating MDIS against well-known AIM saliency method, simulations are needed and described in details with several interesting conclusions, drawn for further research directions.
\end{abstract}

\section{Discriminant Visual Saliency}
\vspace{-5pt}
\label{sec:dis}
Saliency mechanism plays a key role in perceptual organization \cite{treisman1980}; therefore, recently several researchers attempt to generalize principles for visual saliency \cite{itti1998}\cite{bruce2009}\cite{harel2007}\cite{hou2007}\cite{qiu2007},\cite{lengo2012}. In the decision theoretic point of view, saliency is regarded as power for distinguishing salient and non-salient classes; moreover, discriminant saliency, (DIS), combines classical centre-surround hypothesis with derived optimal saliency architecture. Saliency value at a spatial location is identified as the discriminant power of a feature set with respect to the binary classification problem between centre and surround classes. Based on the decision theory, this approach can be generalized for variety of stimulus modalities, including intensity, color, orientation and motion \cite{itti1998}. Moreover, various psychophysical properties for both static and motion stimuli are shown to be accurately satisfied quantitatively by DIS saliency maps \cite{gao2009}. Due to ubiquity of centre-surround operator in the early stages of biological vision, bottom-up saliency is commonly defined as how certain the stimuli at each location of central visual field can be determined against other stimuli in its surround. In other words, ``centre-surround'' hypothesis is also a natural binary classification problem which can be solved by the well-established decision theory. In this problem, classes can be defined as follows.

\begin{itemize}
	\item Centre class: observations within a central neighborhood \(W_{l}^{1}\) of visual fields location \(l\).
	\item Surround class: observations within a surrounding window \(W_{l}^{0}\) of the above central region.
\end{itemize}
Feature responses are drawn from the predefined feature sets \(X\) by a random process. As there are many possible combinations and orders of how such responses are assembled, feature observations can be considered as a random process, \(X(l)=(X_{1}(l),\ldots,X_{d}(l))\) of dimension \(d\).  This random process is drawn conditionally on hidden variable \(Y(l)\) of class states or labels (center / surround). Feature vector \(x(j)\), given \(j\in W_{l}^{c},c\in\{0,1\}\), is drawn from classes \(c\) according to the conditional probability density \(P_{X(l)}\vert_{Y(l)}(x|c)\) where \(Y(l)={0,1}\) are surround and centre labels. The saliency of location l, \(S(l)\) is equal to the discriminant power of X for the classification of the observed feature vectors. That discriminant concept is quantified by mutual information between feature, X and class label, Y.
\begin{linenomath*}
\begin{equation*}
	\begin{aligned}
	S(l) & = I_{l}(X;Y)\\
	 & = \sum_{c}\int p_{X,Y}(x,c)log\frac{p_{X,Y}(x,c)}{p_{X}(x)p_{y}(c)}dx
	\end{aligned}
\end{equation*}
\end{linenomath*}
However, mutual information estimation of \(d\)-dimensional space suffers from the curse of dimensionality. Successfully tackling the problem would make information-based saliency algorithms more biologically plausible and computationally feasible. Dashan Gao and Nuno Vasconcelos have proposed a possible solution called DIS \cite{gao2007}, which is formulated as follows.
\small
\begin{linenomath*}
\begin{equation}
\begin{aligned}
& I_{l}(X;Y) = H(Y)-H(Y\vert X) = \\
& \frac{1}{\vert W_{l}|}\sum_{j\in W_{l}}\left[H(Y)+\sum_{c=0}^{1}P_{Y\vert X}(c\vert x_j)log\, P_{Y\vert X}(c\vert x_j)\right]
\end{aligned}
\label{eq:dis}
\end{equation}
\end{linenomath*}
\normalsize
where \(H(Y)=-\sum_{c=0}^{1}P_{Y}(c)logP_{Y}(c)\) is entropy of classes \(Y\) and \(-E_{Y\vert X}\left[logP_{Y\vert X}(c\vert x)\right]\) is conditional entropy of \(Y\) given \(X\). Given a location l, there are corresponding center \(W_{l}^{1}\) and surround \(W_{l}^{0}\) windows along with a set of associated feature responses \(x(j), j\in W_{l}=W_{l}^{0}\cup W_{l}^{1}\).

While DIS successfully defines discriminant saliency in information-theoretic senses, its implementation, equation \ref{eq:dis}, restrains sampled features in a single fixed-size window. Consequently, it creates a bias toward objects with distinctive features fitted in that window size. As multi-scale processing is an implicit factor of visual attention, DIS needs adapting in wavelet transform, a popular multi-resolution framework.

\section{Multiscale Framework}
\vspace{-5pt}
\label{sec:mdis}
A multi-scale image binary segmentation is a great starting point for multi-scale DIS (MDIS) as it also needs to classify a data point into two classes centre, surround classes. Noted that DIS only uses the binary classification as an intermediate step to measure discriminant value. As segmentation accuracy depends on sizes of classifying windows, an appropriate choice optimizes positive classification ratio; otherwise, it leads to sub-optimal systems. For example, a large window usually provides rich statistical information and enhance reliability of the algorithm; however, it simultaneously risks including heterogeneous elements in the window, which in turn reduces segmentation accuracy.  If processing with too small windows, we probably run into local maxima points while missing global meaningful points. In brief, choosing appropriate window size has vital influence on performances of binary segmentation and consequently of DIS or MDIS.
\subsection{Dyadic Classification Windows}
\label{subsec:mcw}
Dynamic windows with varying sizes can be employed to obtain coarse-to-fine segmented regions \cite{choi1998}. Adapting this approach, MDIS can produce saliency maps with varying resolutions. In MDIS, multiscale dyadic windows are implemented due to its compact arrangement \cite{burt1983}; for example, an initial square image \(s\) with \(2^J\)x\(2^J\) of \(n:=2^{2J}\) pixels, the dyadic square structures can be generated by recursively dividing \(x\) into four square sub-images equally, the left-hand side of figure \ref{fig:quadtree}. Moreover, it is similar to the popular quad-tree structure, commonly employed in wavelet transforms, the right-hand side of figure \ref{fig:quadtree}. Each node of a quad-tree is a child of a node at the directly above level; meanwhile it is a parent of other nodes at the directly below level. Each node corresponds to a dyadic block, combining wavelet coefficients across different sub-bands, nodes \(\tau\) in the figure \ref{fig:quadtree}. Let's denote each block by \(d_{i}^{j}\) given \(i,j\) are indexes of locations, levels. 
\begin{linenomath*}
\begin{figure}[!htbp]%
\centering
\includegraphics[width=0.5\textwidth]{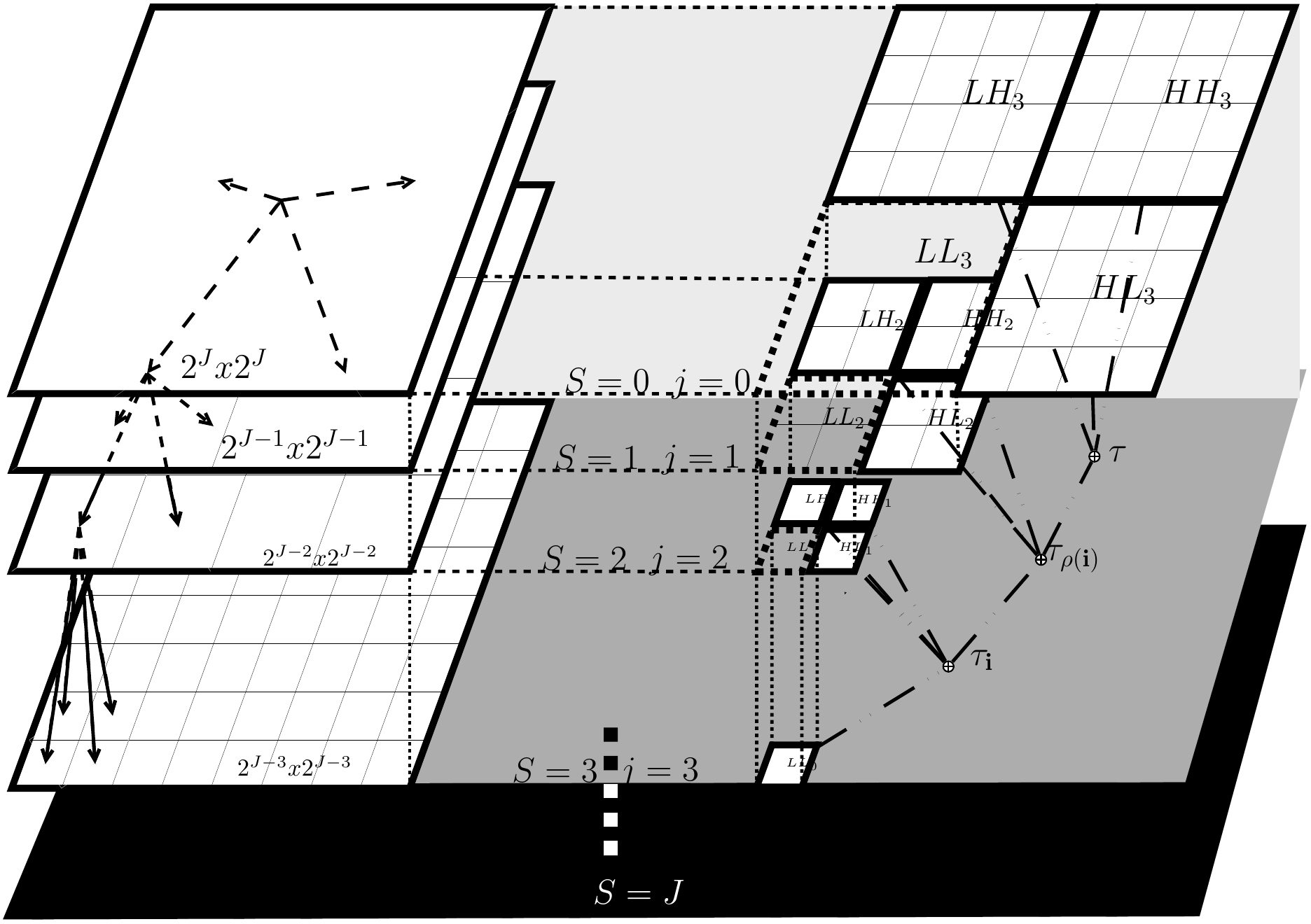}%
\caption{Quad-tree structure}%
\label{fig:quadtree}%
\end{figure}
\end{linenomath*}

Assumed image contents are generated by random variable \(X\), each node of the quad-tree also relates to a randomly generated block. Classification of a node into either centre or surround class requires studying its statistical property. As a node can be represented by wavelet coefficients, Gaussian Mixture Model (GMM) is utilised for estimating their likelihood from mixtures of large and small variance Gaussian distributions. Moreover, inter-scale correlation is usually found between wavelet-coefficients of different levels; hence, this statistical dependence is modelled by Hidden Markov Tree (HMT). Basically, HMT estimates likelihood of each wavelet coefficient give a hidden state, considering feature probability by GSM and transition probability matrix. Noted that, it includes novelty and persistence elements, for which hidden states are probably changed or persisted from open scale to another. Utilization of the up-down algorithm \cite{crouse1998} estimates likelihood ,\( p(d_i^j|c_m) \), of all nodes given their hidden states \( c_m = {0,1} \). Though binary segmentation / classification can be achieved with the maximum likelihood principle, however the results are not consistent across scales due to lack of prior information integration. Choi .et .al \cite{choi2001} proposes a Bayesian Maximum a Posterior (MAP) approach for \( p(c_m|d_i^j,v_i^{j-1}) \), the equation \ref{eq:map5}, whereof both parents' classes and children's features are involved in class decisions. To optimize MAP and enhance across-scale coherency, sweeping operations fuse likelihoods \(f(\mathbf{d_i} \vert c_i)\) along the quad-tree given the label tree prior \(p(c_{i}^{j} \vert \mathbf{v_i})\).
\begin{linenomath*}
\begin{equation}
	\hat{c_i}^{MAP} = \text{argmax}_{c_{i}^{j} \in {0,1}} f(c_{i}^{j} \vert \mathbf{d^j,v^j})
\label{eq:map5}
\end{equation}
\end{linenomath*}
\subsection{Multiscale Discriminant Saliency}
\label{subsec:mdis}
 The DIS method also uses MAP to estimate the scale parameter or variance of GGD (see section 2.4 \cite{gao2009} for more details) as follows.
\begin{linenomath*}
\begin{equation}
\hat{\alpha}^{MAP} = \left[ \frac{1}{{\cal K}} \left( \sum_{j=1}^{n} \vert x(j) \vert^\beta + \nu \right) \right]^{\frac{1}{\beta}}
\label{eq:mdis1}
\end{equation}
\end{linenomath*}
The estimation is later included in centre / surround class decision, the equation \ref{eq:dis}. Therefore, discriminant power is strictly proportional to how difference there are between MAP values of distributions with variances \(\alpha_0, \alpha_1\) from both classes. In MDIS, posterior can be computed directly by the equation \ref{eq:map5}, and its combination with mutual information principle of DIS, the equation \ref{eq:dis} yields a multiscale estimation for discriminant power, \(I_{i}^{j} (C^{j};\mathbf{D^{j}})\).
\begin{linenomath*}
\begin{equation}
H(C^j)+\sum_{c=0}^{1} P_{C^j \vert \mathbf{D^j}}(c_{i}^{j} \vert \mathbf{d^j}) log P_{C^j \vert \mathbf{D^j}}(c_{i}^{j} \vert \mathbf{d^j})
\label{eq:mdis2}
\end{equation}
\end{linenomath*}
Since the equation \ref{eq:mdis2} yields discriminant power across scales, we can choose the maximum MAP values, \(\text{argmax}_{j} \left( I_{i}^{j} \right) \), for each location.
\section{Experiments \& Discussion}
\vspace{-5pt}
\label{sec:exp}
In our paper, we try MAP estimations with several HMT derivatives such as Universal HMT \cite{romberg2001}, Trained HMT \cite{crouse1998}, or Vector HMT \cite{do2002}. Normal HMT (THMT) requires an on-line training stage for estimating model parameters. THMT processes three wavelet orientations independently by single-variate operations; meanwhile, a vector of coefficients can be treated as multi-variate variables in similar operations by VHMT. Multivariate nature of VHMT prefers modelling textural, especially rotation-invariant features. Though THMT or VHMT needs training stages for parameter, they could be fixed by off-line training in UHMT if general image contents are known in advance. Romberg et. al. \cite{romberg2001} have proposed a set of UHMT parameters for natural images, such approach needs evaluating against an established saliency method AIM (An Inforax Method \cite{bruce2006}) in both quantitative (LCC,NSS,AUC,TIME \cite{borji2012-c}) or qualitative measures, visual inspection of generated saliency maps on the well-known Neil Bruce's database \cite{bruce2007-a} with eye-tracking locations.

In the simulation, we deploy five dyadic scales corresponding to (U/T/V)HMT(1-5) of MDIS and integrated saliency maps are denoted by (U/T/V)HMT0. Three numerical measures linear cross correlation (LCC), normalized scanpath saliency (NSS), area under curve AUC and TIME are represented in tables \ref{tab:uhmt}, \ref{tab:thmt}, \ref{tab:vhmt} for (U,T,V)HMT consequently. 
\begin{linenomath*}
\begin{figure*}[!ht]
	\centering
	\begin{subfigure}[b]{0.19\textwidth} 
		\includegraphics[width=\textwidth,height=0.1\textheight]{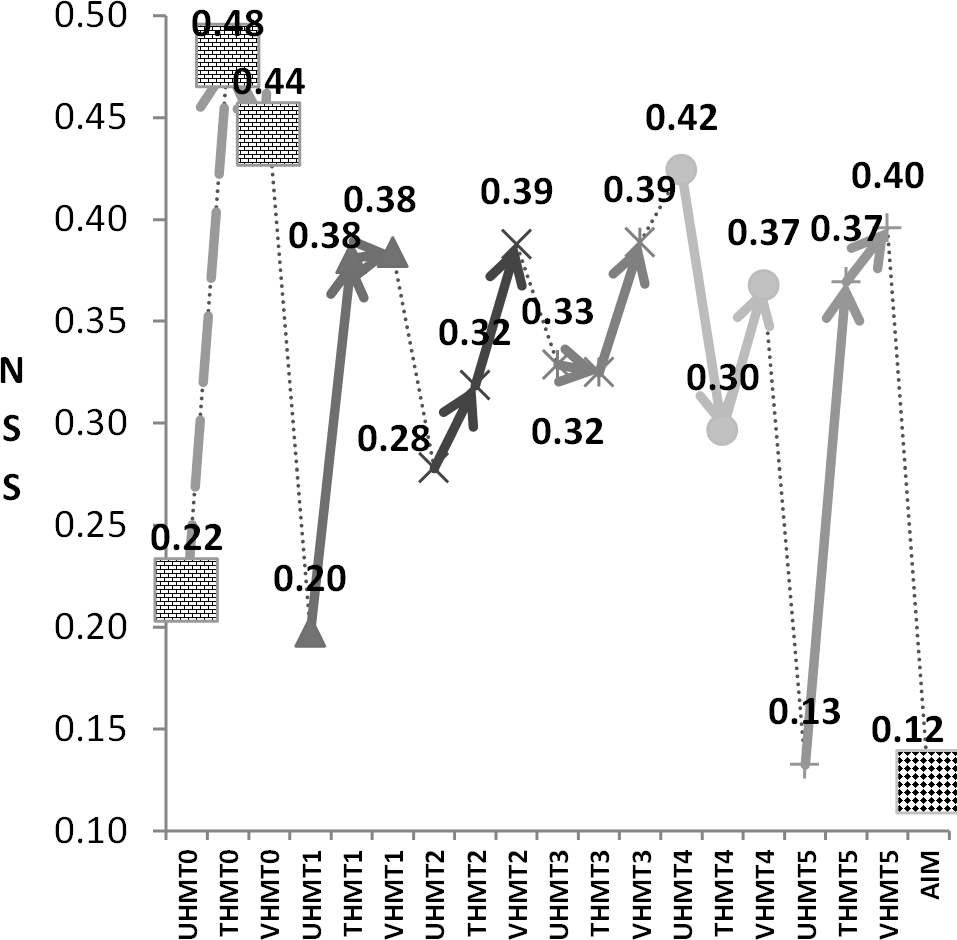}
		\caption{NSS}
		\label{fig:nss}
	\end{subfigure}
	\begin{subfigure}[b]{0.19\textwidth} 
		\includegraphics[width=\textwidth,height=0.1\textheight]{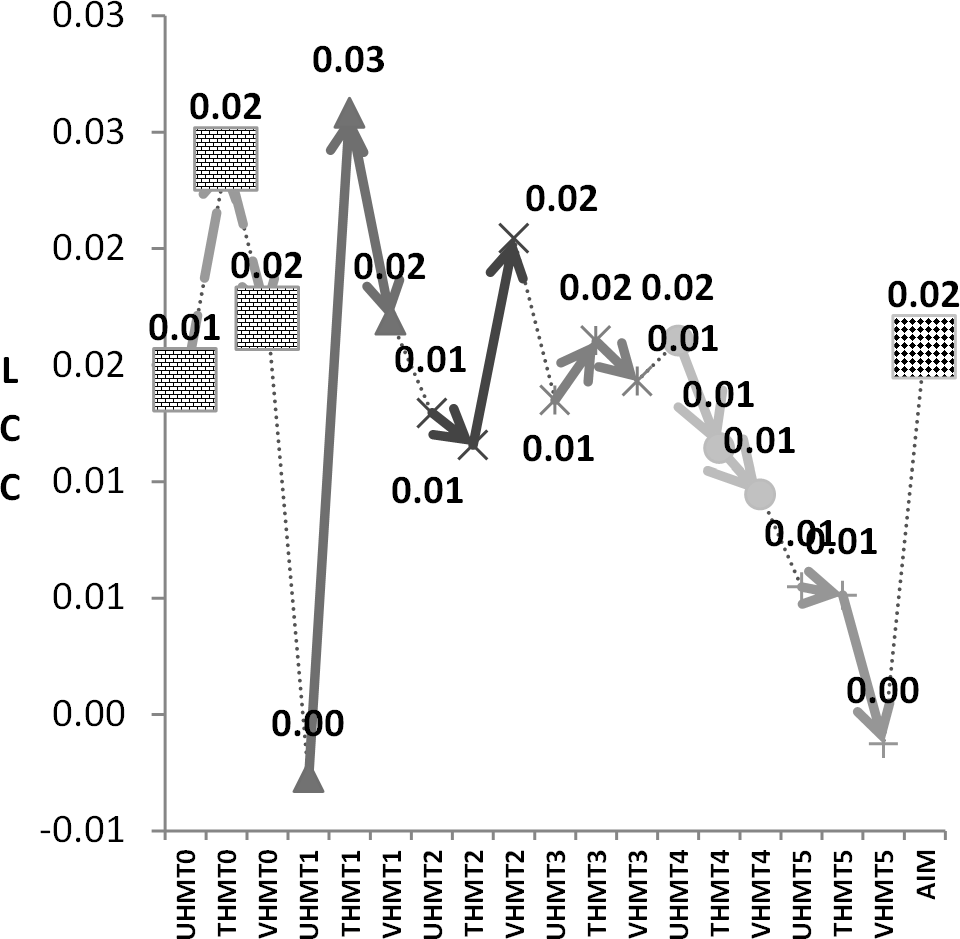}
		\caption{LCC}
		\label{fig:lcc}
	\end{subfigure}
	\begin{subfigure}[b]{0.19\textwidth} 
		\includegraphics[width=\textwidth,height=0.1\textheight]{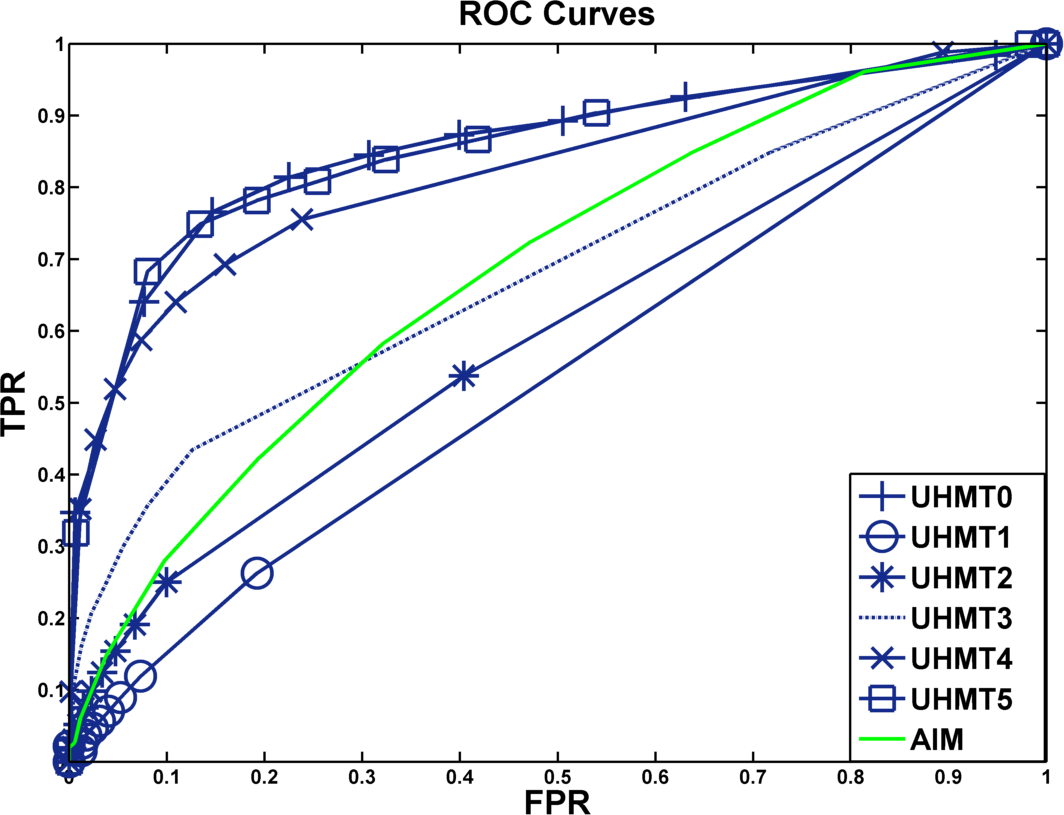}
		\caption{UHMT - ROC}
		\label{fig:uhmt_roc}
	\end{subfigure}	
	\begin{subfigure}[b]{0.19\textwidth} 
		\includegraphics[width=\textwidth,height=0.1\textheight]{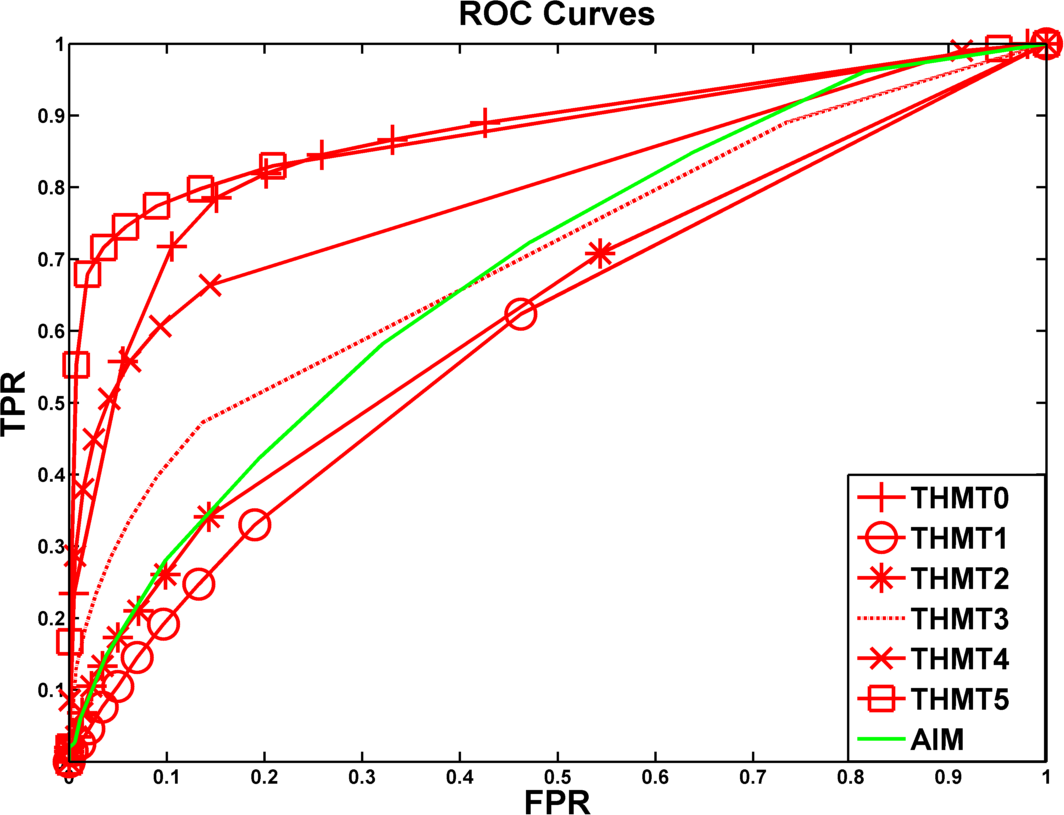}
		\caption{THMT - ROC}
		\label{fig:thmt_roc}
	\end{subfigure}	
	\begin{subfigure}[b]{0.19\textwidth} 
		\includegraphics[width=\textwidth,height=0.1\textheight]{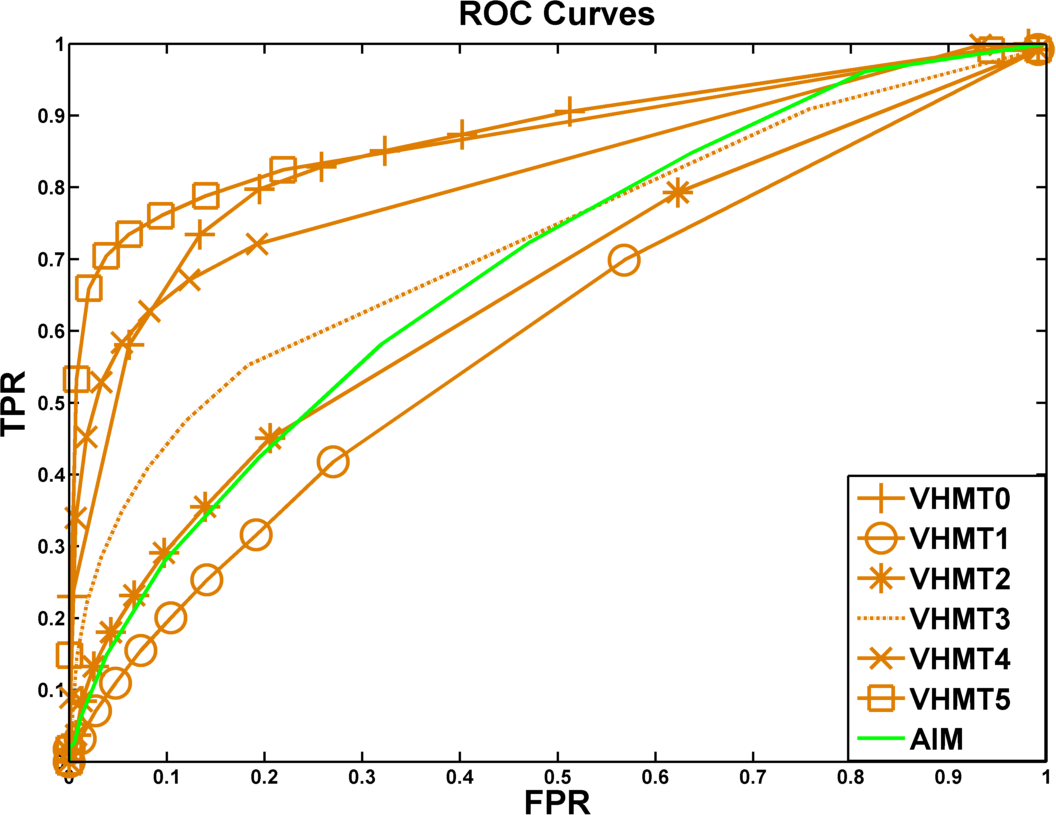}
		\caption{VHMT - ROC}
		\label{fig:vhmt_roc}
	\end{subfigure}				
	\\
	\begin{subfigure}[b]{0.19\textwidth} 
		\includegraphics[width=\textwidth,height=0.1\textheight]{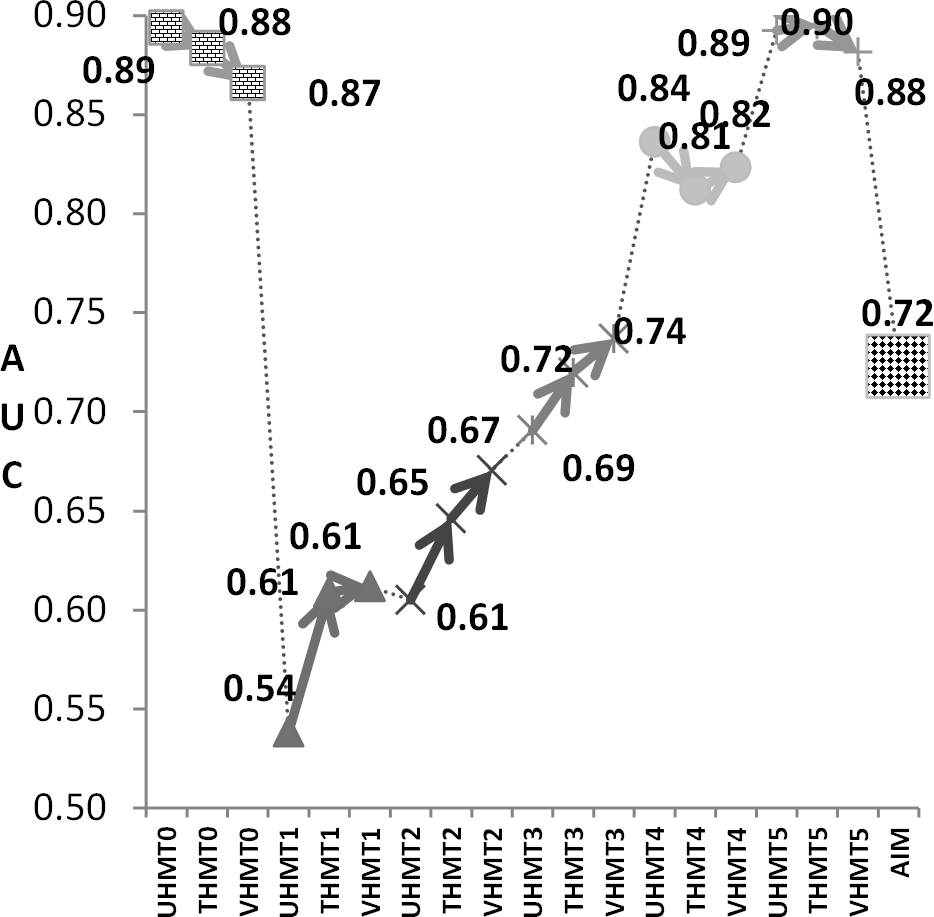}
		\caption{AUC}
		\label{fig:auc}
	\end{subfigure}
	\begin{subfigure}[b]{0.19\textwidth} 
		\includegraphics[width=\textwidth,height=0.1\textheight]{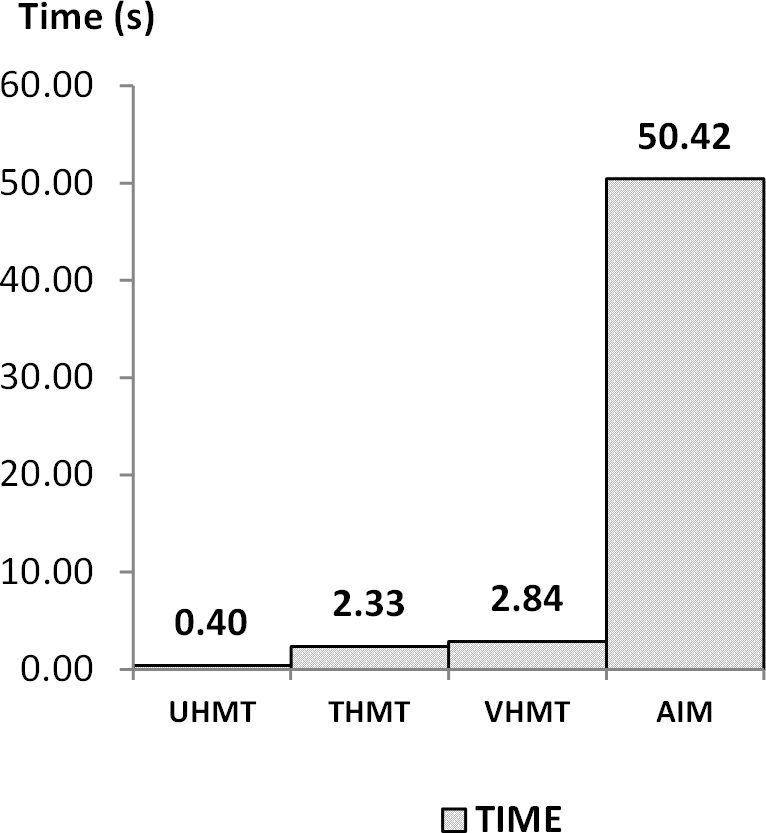}
		\caption{TIME}
		\label{fig:time}
	\end{subfigure}	
	\begin{subfigure}[b]{0.19\textwidth} 
		\includegraphics[width=\textwidth,height=0.1\textheight]{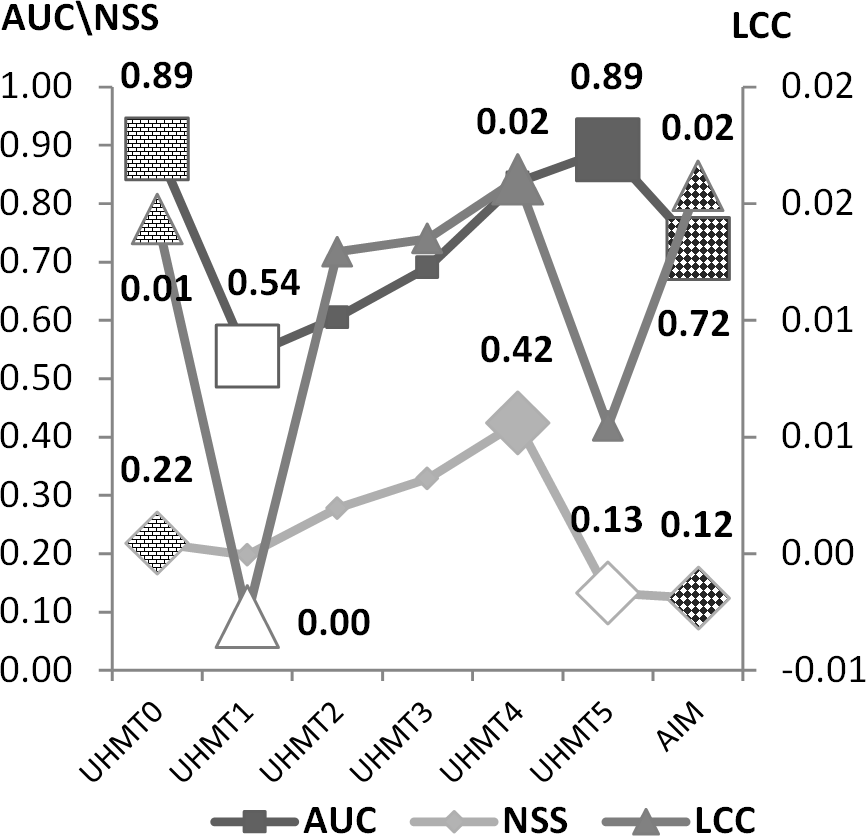}
		\caption{UHMT - MDIS}
		\label{fig:uhmt}
	\end{subfigure}
	\begin{subfigure}[b]{0.19\textwidth} 
		\includegraphics[width=\textwidth,height=0.1\textheight]{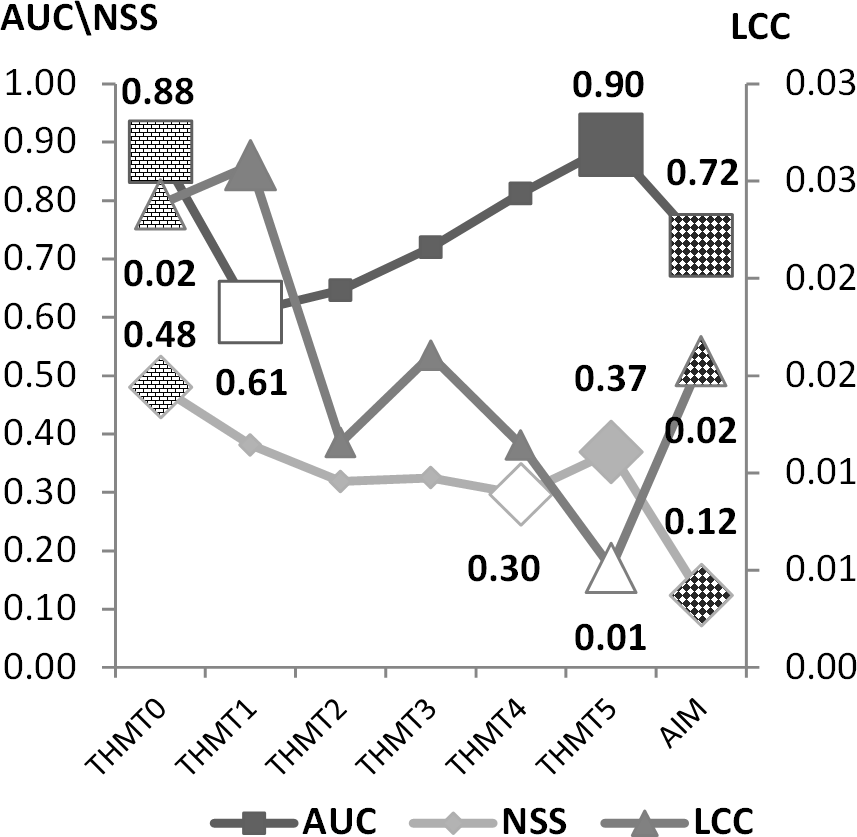}
		\caption{THMT - MDIS}
		\label{fig:thmt}
	\end{subfigure}	
	\begin{subfigure}[b]{0.19\textwidth} 
		\includegraphics[width=\textwidth,height=0.1\textheight]{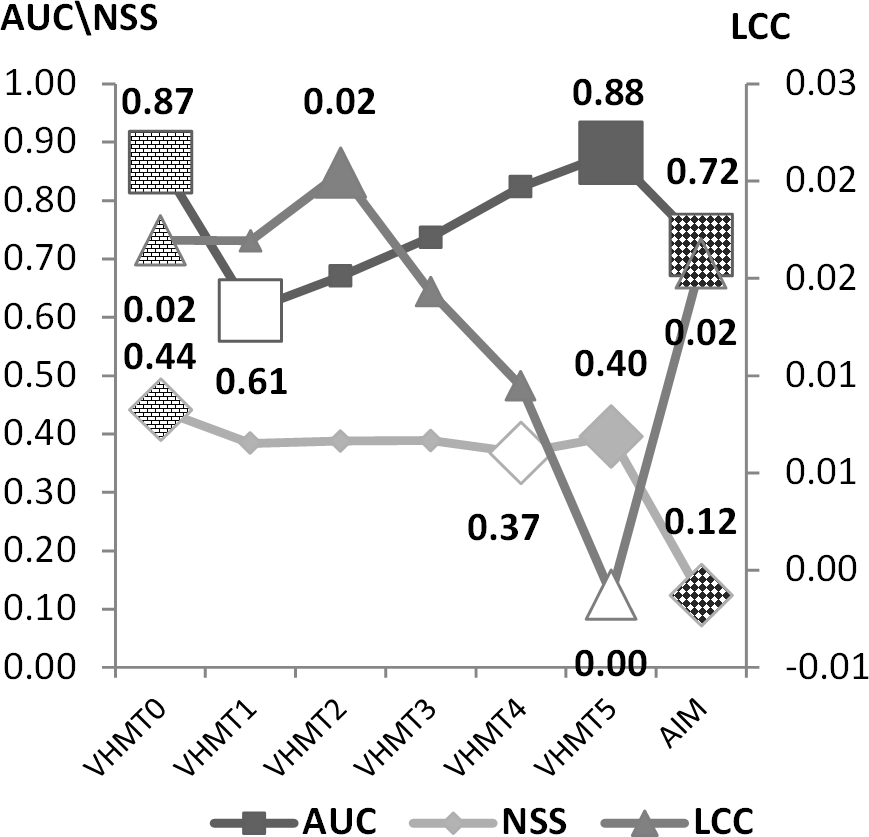}
		\caption{VHMT - MDIS}
		\label{fig:vhmt}
	\end{subfigure}			
	\\
	\begin{subtable}[b]{0.49\textwidth}
		\resizebox{\textwidth}{0.06\textheight}{
		\begin{tabular}{|l|c|c|c|c|}
		\hline
		Observations & LCC & NSS & AUC & TIM \\ \hline
		UHMT0 & 0.01434 & 0.21811 & \textbf{0.89392} & 0.39617 \\ \hline
		UHMT1 & \texttt{-0.00269} & 0.19772 & \texttt{0.53862} & 0.39617 \\ \hline
		UHMT2 & 0.01294 & 0.27819 & 0.60520 & 0.39617 \\ \hline
		UHMT3 & 0.01349 & 0.32868 & 0.69065 & 0.39617 \\ \hline
		UHMT4 & \textbf{0.01604} & \textbf{0.42419} & 0.83615 & 0.39617 \\ \hline
		UHMT5 & 0.00548 & \texttt{0.13273} & \textbf{0.89234} & 0.39706 \\ \hline
		AIM & 0.01576 & 0.12378 & 0.72275 & 50.41714 \\ \hline
		\end{tabular}
		}	
	\caption{UHMT - MDIS - DATA} \label{tab:uhmt}	
	\end{subtable}
	\begin{subfigure}[b]{0.49\textwidth} 
		\includegraphics[width=\textwidth,height=0.12\textheight]{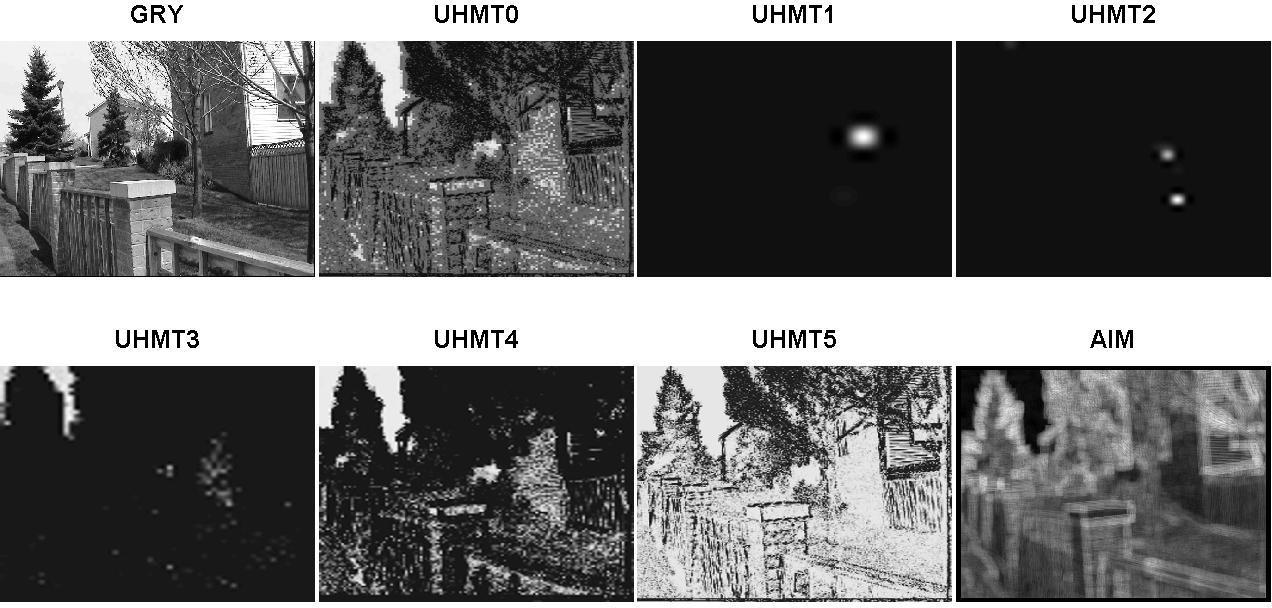}
		\caption{UHMT - MDIS - MAP}
		\label{fig:uhmt_102}
	\end{subfigure}	
	\\
	\begin{subtable}[b]{0.49\textwidth}
		\resizebox{\textwidth}{0.06\textheight}{
		\begin{tabular}{|l|c|c|c|c|}
		\hline
		Observations & LCC & NSS & AUC & TIM \\ \hline
		THMT0 & \textbf{0.02382} & \textbf{0.48019} & \textbf{0.88357} & 2.32734 \\ \hline
		THMT1 & \textbf{0.02582} & 0.38096 & \texttt{0.60922} & 2.32734 \\ \hline
		THMT2 & 0.01156 & 0.31855 & 0.64633 & 2.32726 \\ \hline
		THMT3 & 0.01604 & 0.32491 & 0.71972 & 2.32726 \\ \hline
		THMT4 & 0.01143 & \texttt{0.29662} & 0.81192 & 2.32726 \\ \hline
		THMT5 & \texttt{0.00512} & 0.36932 & \textbf{0.89532} & 2.32726 \\ \hline
		AIM & 0.01576 & 0.12378 & 0.72353 & 50.41714 \\ \hline
		\end{tabular}
		}	
	\caption{THMT - MDIS - DATA} \label{tab:thmt}	
	\end{subtable}
	\begin{subfigure}[b]{0.49\textwidth} 
		\includegraphics[width=\textwidth,height=0.12\textheight]{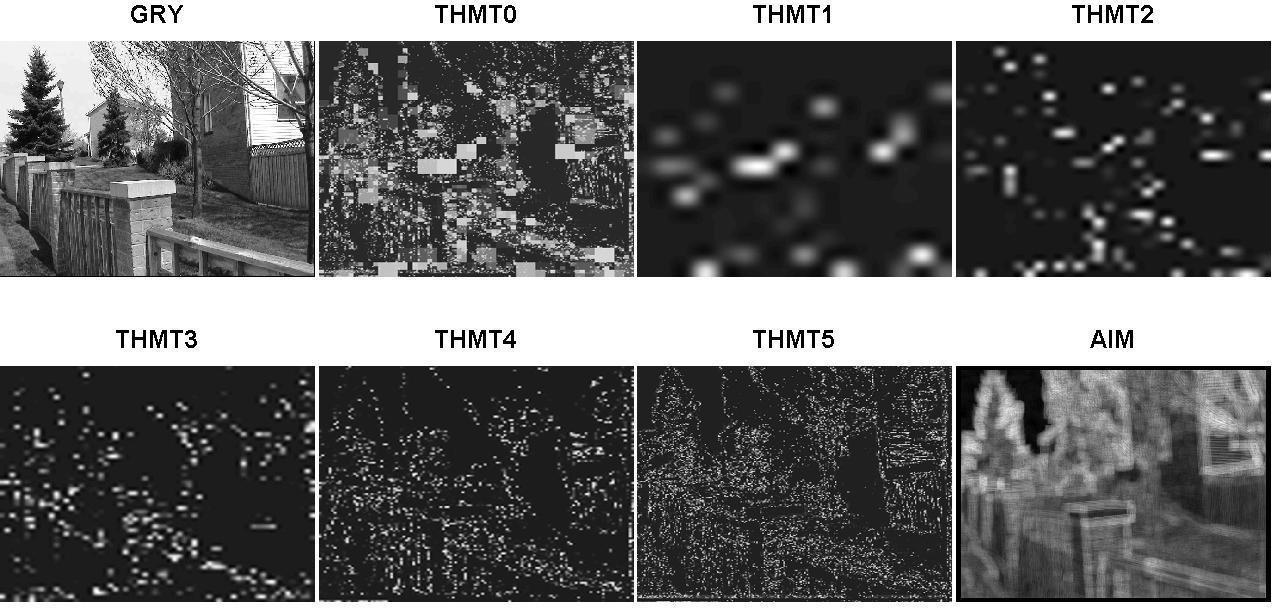}
		\caption{THMT - MDIS - MAP}
		\label{fig:thmt_102}
	\end{subfigure}		
	\\
	\begin{subtable}[b]{0.49\textwidth}
		\resizebox{\textwidth}{0.06\textheight}{
		\begin{tabular}{|l|c|c|c|c|}
		\hline
		Observations & LCC & NSS & AUC & TIM \\ \hline
		VHMT0 & 0.01697 & \textbf{0.44170} & \textbf{0.86606} & 2.84212 \\ \hline
		VHMT1 & 0.01693 & 0.38387 & \texttt{0.61187} & 2.84212 \\ \hline
		VHMT2 & \textbf{0.02044} & 0.38777 & 0.67060 & 2.84212 \\ \hline
		VHMT3 & 0.01430 & 0.38882 & 0.73682 & 2.84212 \\ \hline
		VHMT4 & 0.00946 & \texttt{0.36761} & 0.82329 & 2.84212 \\ \hline
		VHMT5 & \texttt{-0.00125} & 0.39580 & \textbf{0.88160} & 2.84212 \\ \hline
		AIM & 0.01576 & 0.12378 & 0.72400 & 50.41714 \\ \hline
		\end{tabular}
		}	
	\caption{VHMT - MDIS - DATA} \label{tab:vhmt}	
	\end{subtable}
	\begin{subfigure}[b]{0.49\textwidth} 
		\includegraphics[width=\textwidth,height=0.12\textheight]{thmt_102.jpg}
		\caption{VHMT - MDIS - MAP}
		\label{fig:vhmt_102}
	\end{subfigure}		
    \captionlistentry[table]{Quantitative and Qualitative evaluation of MDIS and AIM}
    \captionsetup{labelformat=andtable}
    \caption{Quantitative and Qualitative evaluation of MDIS and AIM}	
\end{figure*}
\end{linenomath*}
In these tables, TIME represents computational requirement of saliency methods of (U,T,V) HMT which are listed in predictable incrementing orders. While UHMT requires the least TIME due to no requirement for training, THMT and VHMT need more computational effort for learning model parameters in single and multiple variate manners . (T,V)HMT surpass UHMT in evaluated LCC, NSS, and AUC scores, shown in the tables \ref{tab:uhmt},\ref{tab:thmt},\ref{tab:vhmt} and figures \ref{fig:nss},\ref{fig:auc},\ref{fig:lcc}. Comparatively, the proposed MDIS surpasses AIM in all quantitative measures, clearly shown by each column of these tables with \textbf{maximum} and \texttt{minimum} values. In figures \ref{fig:uhmt_roc},\ref{fig:thmt_roc},\ref{fig:vhmt_roc} are shown the comparisons between different modes of MDIS and AIM with Receiver Operating Curve (ROC). Generally, HMT-based MDIS modes perform better than AIM in smaller scales (U,T,V)HMT(0,4,5) but MDIS in larger scales HMT(1,2,3) are equivalent or slight worse than AIM. AUC measures are increased with shrinking sizes of processing windows HMT(1-5) regardless of U/T/V modes. Meanwhile, LCC and NSS are varied more wildly, for instance, UHMT has the best LCC, NSS at the HMT4 mode; while, (T,V)HMT almost has the best evaluation at HMT0, the integrated mode. Overall, trained HMT, especially VHMT in the table \ref{tab:vhmt} and figure \ref{fig:vhmt}, provides more consistent numerical results through different scales. Figures \ref{fig:uhmt_102},\ref{fig:thmt_102},\ref{fig:vhmt_102} show sample saliency maps of (U,T,V)HMT(0-5) MDISs and AIM for qualitative evaluation.  (T,V)HMT have similar saliency maps while the UHMT map highlights unlikely attentive regions. Its poor performance might be due to lack of training steps.
\section{Conclusion}
\vspace{-5pt}
\label{sec:cls}
In conclusion, Multiscale Discriminant Saliency (MDIS) is developed as an extension of DIS \cite{gao2007-a} under the dyadic scale framework of wavelet transform. MDIS utilizes mutual information between classes and feature distribution to quantify classifying discriminant power as saliency value in multiple dyadic-scale structures. Moreover, it fuses prior information, class decisions from previous scales, in Bayesian MAP along quad-tree in coarse-to-fine manner to create consistent saliency maps for multiple scales and final integrated map with maximum information rule. MDISs are evaluated against AIM to prove MDIS's competitiveness. For further research direction is implementation of MDIS algorithms on embedded and mobile platforms.
\ifisieee
\bibliographystyle{IEEEbib}
\fi
\bibliography{../BibRefs/refs.bib}
\end{document}